\documentclass[10pt,twocolumn,letterpaper]{article}

\usepackage[pagenumbers]{cvpr} 

\usepackage[dvipsnames]{xcolor}

\definecolor{cvprblue}{rgb}{0.21,0.49,0.74}
\usepackage[pagebackref,breaklinks,colorlinks,citecolor=cvprblue]{hyperref}
\usepackage{makecell}
\usepackage{multirow}
\usepackage{CJKutf8}
\usepackage{amsmath}
\usepackage{indentfirst}

\def\paperID{8932}
\def\confName{CVPR}
\def\confYear{2024}

\title{RCooper: A Real-world Large-scale Dataset for \\ Roadside Cooperative Perception}

\author{Ruiyang Hao$^{1,\dag}$, Siqi Fan$^{1,\dag}$, Yingru Dai$^{1,2}$, Zhenlin Zhang$^{3}$, Chenxi Li$^{3}$, Yuntian Wang$^{3}$, \\ Haibao Yu$^{1,4}$, Wenxian Yang$^{1}$, Jirui Yuan$^{1}$, Zaiqing Nie$^{1,}$\thanks{Corresponding author. \dag\,
indicates equal contribution. Work done at AIR. For any questions, please email dair@air.tsinghua.edu.cn.} \\
$^{1}$Institute for AI Industry Research (AIR), Tsinghua University \\
$^{2}$ Department of Electronic Engineering, Tsinghua University \\ $^{3}$ China Automotive Innovation Corporation \quad $^{4}$ The University of Hong Kong
}

\begin{document}
\begin{CJK}{UTF8}{gbsn}
\maketitle
\begin{abstract}
The value of roadside perception, which could extend the boundaries of autonomous driving and traffic management, has gradually become more prominent and acknowledged in recent years. However, existing roadside perception approaches only focus on the single-infrastructure sensor system, which cannot realize a comprehensive understanding of a traffic area because of the limited sensing range and blind spots. Orienting high-quality roadside perception, we need \textbf{R}oadside \textbf{Coo}perative \textbf{Per}ception \textbf{(RCooper)} to achieve practical area-coverage roadside perception for restricted traffic areas. Rcooper has its own domain-specific challenges, but further exploration is hindered due to the lack of datasets. We hence release the first real-world, large-scale RCooper dataset to bloom the research on practical roadside cooperative perception, including detection and tracking. The manually annotated dataset comprises 50k images and 30k point clouds, including two representative traffic scenes (i.e., intersection and corridor). The constructed benchmarks prove the effectiveness of roadside cooperation perception and demonstrate the direction of further research. Codes and dataset can be accessed at: \hyperlink{https://github.com/AIR-THU/DAIR-RCooper}{https://github.com/AIR-THU/DAIR-RCooper}.
\end{abstract}
    
\section{Introduction}
\label{sec:intro}

With the development of the Internet of Things (IoT), 5G, and artificial intelligence technologies, the value of roadside perception has gradually become more prominent, which has drawn broad attention in recent years \cite{rope3d, yu2022dair, bevheight, monogae, cbr, bevheight++}. Roadside perception is of great significance to both autonomous driving and traffic management. For autonomous driving, roadside sensor systems provide intelligent vehicles with complementary on-road messages beyond the onboard perspective, assisting vehicles to have a more comprehensive and clear understanding of surrounding environments, thereby trending to better and safer driving of L5 autonomous \cite{han2023collaborative}. As for intelligent transportation systems, the downstream traffic management task, e.g., traffic flow control, traffic participants monitoring, and illegal activities monitoring, can be further improved with more comprehensive understanding by roadside perception \cite{zhang2011data}. Therefore, orienting high-quality autonomous driving and traffic management, how to achieve practical area-coverage roadside perception for a restricted traffic area is a significant task.

\begin{figure}[t]
  \centering
  \includegraphics[scale=0.29]{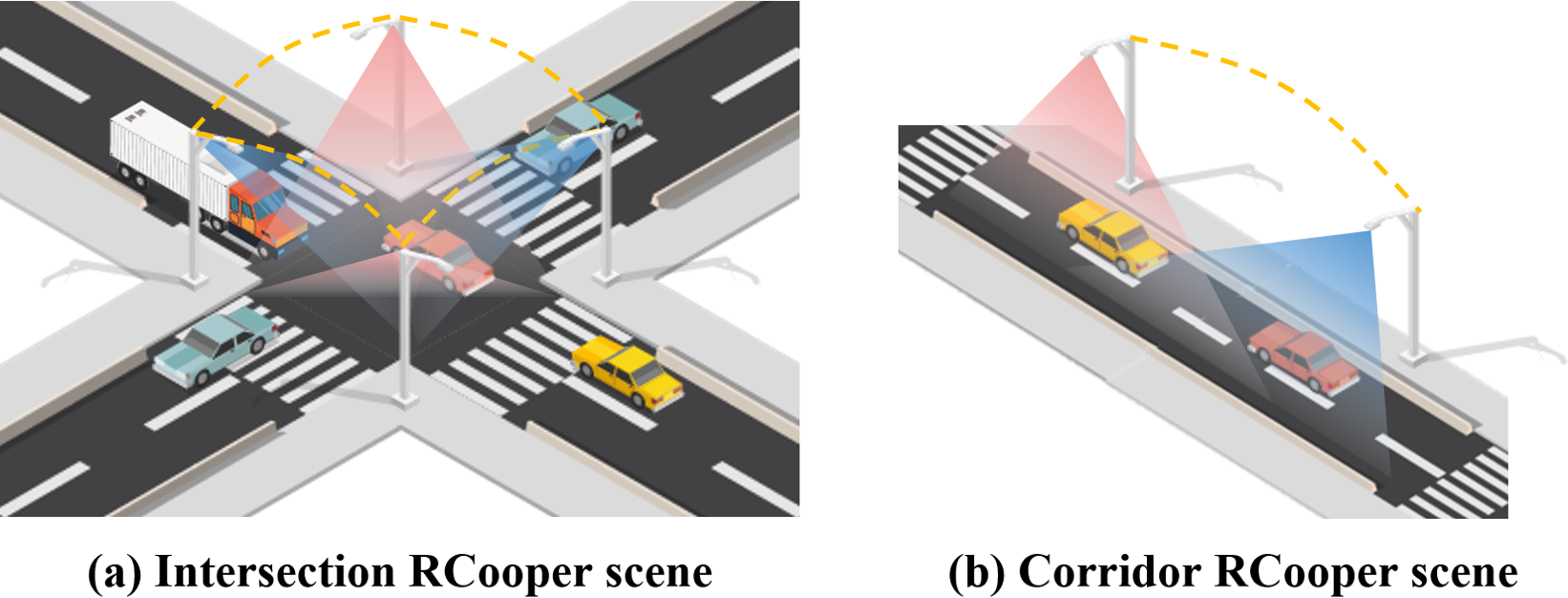}
  \caption{\textbf{R}oadside \textbf{Coo}perative \textbf{Per}ception \textbf{(RCooper)} is expected to achieve practical area-coverage roadside perception for restricted traffic areas, which would further promote both the autonomous driving and traffic management. The complex roadside system is boiled down to two typical roadside settings, i.e., (a) intersection RCooper scenes and (b) corridor RCooper scenes.}
  \label{fig:intro}
\end{figure}

\begin{figure*}[t]
  \centering
  \includegraphics[scale=0.52]{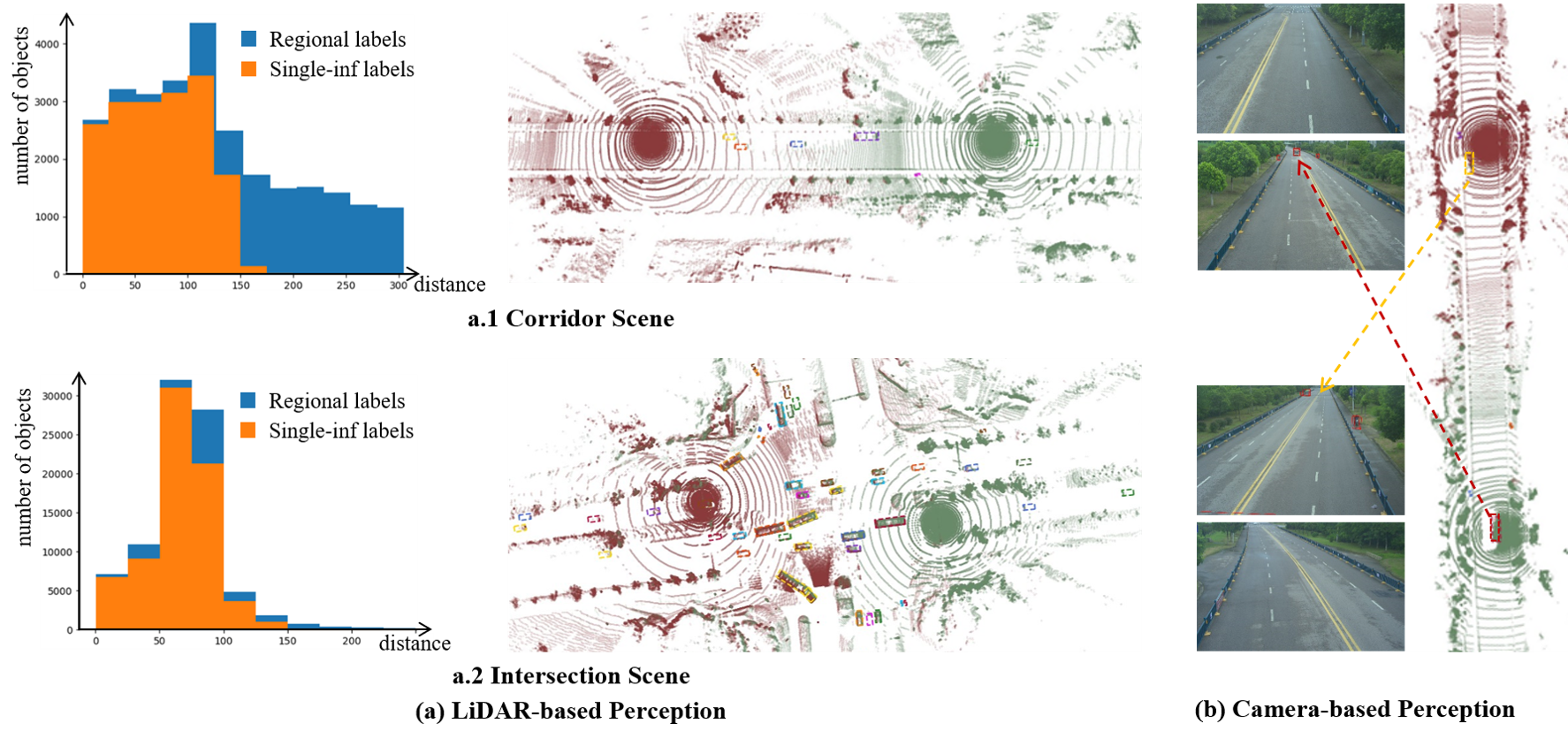}
  \caption{Independent roadside 3D perception (red point clouds) is limited by sensing range and blind spots. (a) The infrastructure-side cooperation can effectively extend the sensing range to cover the whole corridor scene, and the observation from multiple views can weaken the impact of occlusion in the complex intersection scene. (b) The area under the infrastructure is the camera's blind spot, which is perceptible from the adjacent infrastructure's camera.}
  \label{fig:ad-coop}
\end{figure*}

Previous roadside perception \cite{bevheight, bevheight++, monogae, cbr, huang2022rd, zimmer2023tumtraf} mainly concentrates on the perception from an independent roadside view due to the accessible dataset \cite{rope3d, yu2022dair}. However, they can not realize a comprehensive understanding of a traffic area. Single-infrastructure roadside perception is constrained by the installation perspective, leading to the limited sensing range and blind spots, which can be handled via cross-infrastructure cooperation. Observation from various views can extend the sensing range, reduce blind spots, and further enhance the understanding of the same instance. Towards practical 
applications,
\textbf{R}oadside \textbf{Coo}perative \textbf{Per}ception \textbf{(RCooper)} is expected to achieve area-coverage roadside perception for restricted traffic areas, which is shown in Fig.~\ref{fig:intro}. The capabilities of RCooper, including extending the sensing range and reducing blind spots, are illustrated in Fig.~\ref{fig:ad-coop}.

Technically, three challenges can be derived from RCooper: \textit{1) Data heterogeneity}. Considering the construction cost, various types of sensors (multiline LiDAR, MEMS LiDAR, and camera) are employed in practice, leading to prominent data heterogeneity for cooperative perception \cite{bai2023vinet, fang2023lidar}. \textit{2) Cooperative representation merits further enhancement}. Most existing cooperative perception approaches, including well-explored vehicle-vehicle (V2V) and vehicle-infrastructure (V2I) cooperation, are designed for vehicle-centric cooperative tasks \cite{who2com, chen2022model, qiao2023adaptive, zhou2022multi, lei2022latency, vimi, ffnet}. However, to our knowledge, the roadside cooperation approaches have not been investigated before. The inherent characteristics of roadside sensors (like roll, pitch angle, height) make roadside cooperative representation a different playground compared with vehicle-centric cooperation \cite{rope3d, monogae, bevheight, zimmer2023infradet3d}. \textit{3) Perception performance needs improvements}. How to achieve high-quality downstream perception tasks based on roadside cooperative representation, e.g., detection, tracking, counting, and monitoring, needs further investigation. For example, tracking by unstable detection results in complex intersection scenes is still challenging. It is necessary to delve into these challenges, but the lack of datasets hinders the exploration of the Garden of Eden.

We hence release the first real-world, large-scale dataset RCooper for this challenging field to open the gate and boost the development of roadside cooperative perception. We follow the installation scheme widely adopted in practical applications. The complex roadside system is boiled down to two typical roadside settings, i.e., intersection and corridor. We select several representative locations with different levels of traffic flow for each traffic scene, resulting in a manually annotated dataset comprising 50k images and 30k point clouds, which covers diverse weather and lighting variations.

Our contributions are summarized as follows:
\begin{itemize}
  \item The first real-world, large-scale dataset, RCooper, is released to bloom research on roadside cooperative perception for practical applications. All the frames and scenes are captured in real-world scenarios.
  \item More than 50k images and 30k point clouds manually annotated with 3D bounding boxes and trajectories for ten semantic classes are provided in our RCooper, which enables the training and evaluation of roadside cooperative perception approaches in real-world scenarios.
  \item Two cooperative perception tasks, including 3D object detection and tracking, are introduced, and comprehensive benchmarks with SOTA methods are reported. The results show the effectiveness of roadside cooperation and demonstrate the direction of further research.
\end{itemize}

\section{Related work}
\label{sec:relatedwork}

This section briefly reviews three related topics: roadside perception, cooperative perception, and public perception datasets in roadside systems.

\begin{table*}[t]
  \small
  \centering
    \begin{tabular}{p{4.0em}|c|c|c|c|c|c|c|c|c}
    \Xhline{1.2pt}
    \multicolumn{1}{c|}{\textbf{Dataset type}} & \textbf{Source} & \textbf{Dataset} & \textbf{Year}  & \textbf{Coop Mode}  & \textbf{RGBs} & \textbf{LiDARs} & \textbf{Catagories} & \textbf{Det task} & \textbf{Trk task} \\
    \Xhline{1.2pt}
    \multirow{5}[1]{*}{Roadside} & \multirow{5}[1]{*}{real} & BoxCars \cite{boxcars} & 2018  & None   & 116k    & -  & 12    & -    & - \\
    \multicolumn{1}{c|}{} &       & BAAI-VANJEE \cite{baai} & 2021  & None   & 5k    & 2.5k  & 12    & 3D    & - \\
    \multicolumn{1}{c|}{} &       & Rope3D \cite{rope3d} & 2021  & None  & 50k   & 50k   & 13    & 3D    & - \\
    \multicolumn{1}{c|}{} &       & Mona \cite{mona} & 2022  & None     & 11.7M & -     & 2     & 2D    & 2D \\
    \multicolumn{1}{c|}{} &       & A9 \cite{a9} & 2022  & None & 5.4k  & 5.3k  & 10    & 3D    & 3D \\
    \hline
    \multirow{7}[4]{*}{Cooperative} & \multirow{5}[2]{*}{sim} & CODD \cite{COOD} & 2021  & V2V  & -     & 13.5k & 2     & -     & - \\
    \multicolumn{1}{c|}{} &       & OPV2V \cite{xu2022opv2v} & 2022  & V2V  & 44k   & 11k   & 1     & 3D    & - \\
    \multicolumn{1}{c|}{} &       & V2X-Sim \cite{li2022v2x} & 2022  & V2X   & 60k   & 10k   & 1     & 3D    & 3D \\
    \multicolumn{1}{c|}{} &       & V2XSet \cite{xu2022v2x} & 2022  & V2X  & 44k   & 11k   & 1     & 3D    & - \\
    \multicolumn{1}{c|}{} &       & DOLPHINS \cite{dolphins} & 2022  & V2X  & 127k  & 84k   & 3     & 2D \& 3D & - \\
\cline{2-10}    \multicolumn{1}{c|}{} & \multirow{2}[2]{*}{real} & DAIR-V2X \cite{yu2022dair} & 2022  & V2X  & 39k   & 39k   & 10    & 3D    & - \\
    \multicolumn{1}{c|}{} &       & V2V4Real \cite{v2v4real} & 2023  & V2V  & 40k   & 20k   & 5     & 3D    & 3D \\
    \hline
    \multicolumn{1}{m{1.8cm}|}{\textbf{Roadside \quad Cooperative}} & \textbf{real} & \textbf{Rcooper (Ours)} & \textbf{2024} & \textbf{Roadside} & \textbf{50k} & \textbf{30k} & \textbf{10} & \textbf{3D} & \textbf{3D} \\
    \Xhline{1.2pt}
    \end{tabular}
  \caption{Comparisons among the representative public perception dataset for road systems.}
  \label{tab:datasets_com}
\end{table*}

\subsection{Roadside Perception}

Benefiting from the release of roadside public datasets, such as Rope3D \cite{rope3d} and DAIR-V2X-I \cite{yu2022dair}, several pioneer roadside perception methods have emerged in recent years. A simple and effective attempt utilizing camera specifications and the ground knowledge is proposed along with the dataset in Rope3D \cite{rope3d}. MonoGAE \cite{monogae} further proposes a ground-aware embedding to integrate implicit roadside ground information with high-dimensional semantic features. BEVHeight \cite{bevheight} discovers the importance of predicting the height to the ground to ease the optimization process of camera-based roadside perception, and the follow-up work BEVHeight++ \cite{bevheight++} further enhances the performance by fusing the height and depth representation. Considering the practical challenges of calibration noises, CBR \cite{cbr} achieves calibration-free roadside perception via decoupled feature reconstruction. Limited by the available dataset, existing methods do their utmost to pursue better and more robust perception performance with independent roadside sensor systems. However, `two \textit{eyes} are better than one,' we believe the cross-infrastructure cooperation can further boost the roadside perception performance.

\subsection{Cooperative Perception}
According to the collaboration stage, cooperative perception can be divided into early, intermediate, and late fusion \cite{9732063, bai2022survey, han2023collaborative}. Since the 3D point cloud has inherent aggregation convenience, early fusion approaches usually adopt LiDAR as a sensor \cite{cooper, 9228884}, and collaboration on raw data with comprehensive information makes it always become the upper bound of cooperative perception. However, the massive amount of data also introduces high transmission and computation costs. On the other hand, late fusion is bandwidth-economic for only transferring perception results \cite{9228884, yu2022dair}. The fusion strategy is physically explicable but relies on accurate individual predictions. Recent methods focus more on intermediate fusion to balance the trade-off between performance and cost. They focus on either fusion strategy for better performance \cite{wang2020v2vnet, li2021learning, xu2022v2x, when2com, cui2022coopernaut, fcooper, coca3d, cobevt}, or feature selection for transmission efficiency \cite{where2comm, 9682601}. Unlike scene-level feature cooperation, instance-level query cooperation is proposed for interpretable flexible feature interaction \cite{quest}. Compared with well-explored vehicle-centric cooperative perception, the potential of roadside systems has yet to be fully exploited. We introduce a new roadside cooperative perception playground called RCooper.

\subsection{Perception Datasets in Road System}
The blooming development of data-driven perception for autonomous driving and traffic management has dramatically benefited from abundant public traffic scene datasets. The pioneering work, KITTI \cite{kitti}, released the first well-known dataset for autonomous driving, and nuScenes \cite{nuscenes} provided $360^{\circ}$ view multimodal data to boost single-vehicle perception research further. To fuel the development in cooperative perception, various multi-agent datasets have emerged in recent years, but most of them are derived from simulators (e.g., CARLA \cite{carla} and OpenCDA \cite{opencda}) due to the difficulty of collecting actual data \cite{COOD, xu2022opv2v, 9228884, xu2022v2x, dolphins, li2022v2x}. DAIR-V2X \cite{yu2022dair, v2x-seq} and V2V4Real \cite{v2v4real} are the two large-scale real-world datasets for vehicle-centric cooperative perception, which is significant to practical application in natural scenes. Roadside perception has drawn more attention for its comprehensive perception capability, and several single-infrastructure datasets \cite{boxcars, baai, rope3d, mona, a9} are publicly available. However, there is a lack of public datasets for roadside cooperative perception. To facilitate the exploration of this exciting and challenging field, we release the first real-world, large-scale dataset, RCooper, in this paper. The comparisons among the representative public dataset for perception in road systems are reported in Tab.~\ref{tab:datasets_com}.
\section{RCooper Dataset}
\label{sec:dataset}


To advance roadside cooperative perception, we introduce \textbf{RCooper}, a real-world, large-scale, multi-modal dataset annotated with 3D bounding boxes and trajectories. We commence with the data acquisition methodology, delineate the annotation process, and present data analysis.

\subsection{Data Acquisition}

\begin{figure}[t]
  \centering
  \includegraphics[scale=0.4]{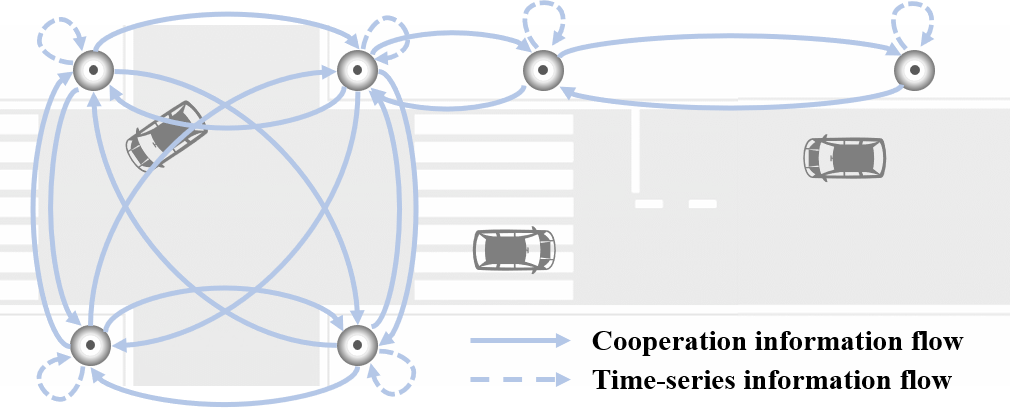}
  \caption{RCooper scenes can be modeled as the structural-stable graph, where the infrastructures and connections are regarded as nodes and edges. Two basic units in graphics, line segment and loop, correspond to the corridor and intersection in actual scenes.}
  \label{fig:scene_modeling}
\end{figure}

\paragraph{Scenario Selection} Figure~\ref{fig:scene_modeling} depicts the road network as a graph with line segments and loops representing corridors and intersections, respectively. These form the basis for two primary traffic scene types. Besides, the dataset is enriched by capturing scenarios across various times and weather conditions throughout the year, ensuring a wide range of environmental and lighting diversities.

\begin{figure}[t]
  \centering
  \includegraphics[scale=0.33]{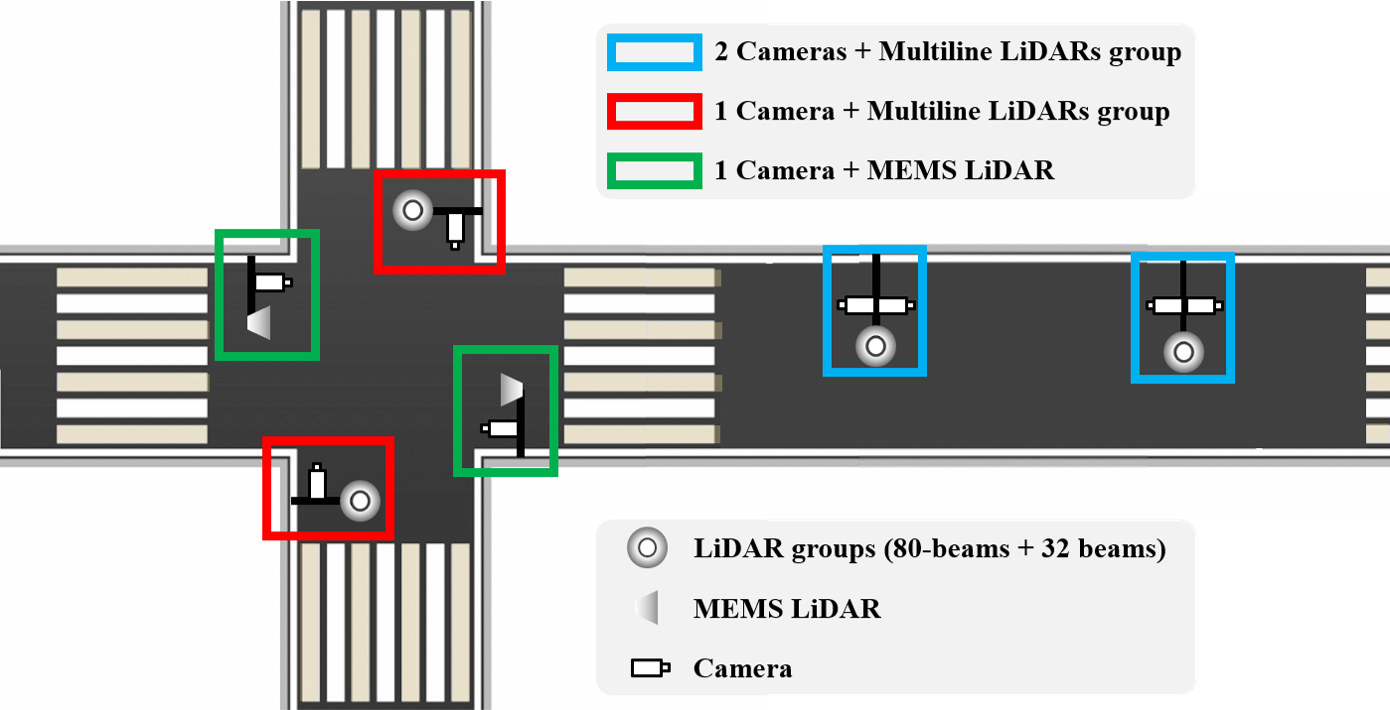}
  \caption{Diagram of the infrastructure-side sensor system. For intersection scenes, we adopt a hybrid scheme for LiDAR-based systems where 2 LiDARs groups (80-beams + 32-beams) and 2 MEMS LiDAR are utilized, because the integration of multiline and MEMS LiDAR is gaining traction for its cost-effectiveness compared to using multiline LiDAR alone. For corridor scenes, each sensor agent include a LiDARs group to cover the region.}
  \label{fig:install}
\end{figure}

\paragraph{Sensor System Design} We follow the typical installation scheme of the infrastructure-side sensor system in practical applications, as shown in Fig.\ref{fig:install}. Considering the characteristics of scenes and construction cost, there are three schemes for the infrastructure agents. Different from the vehicle-side sensor system, 2 LiDARs with different beams (80 beams + 32 beams) are combined as a group for roadside systems since the mounting height makes a single LiDAR can not sense the area directly below it. Three specific installation schemes are illustrated below, shown in Fig.\ref{fig:install}.
\begin{itemize}
  \item \textbf{2 Cameras + Multiline LiDARs Group} is adopted in corridor scenes. The corridor area is long and narrow, which is difficult for a single agent to cover. Fig.~\ref{fig:ad-coop} (a.1) shows that the sensing area of the two neighboring LiDAR systems is intersected, which achieves full area coverage. 2 cameras on the same agent are mounted in reverse, and the cameras of neighboring agents can capture the blind spot directly below itself (Fig.~\ref{fig:ad-coop} (b)). 
  \item \textbf{1 Camera + Multiline LiDARs Group} is adopted in intersection scenes. The camera is mounted towards the junction to capture the RGB video, while the multiline LiDAR can sense half the area of the scene. Such 2 agents are placed in opposite to cover the most of area.
  \item \textbf{1 Camera + MEMS LiDAR} is also adopted in intersection scenes to achieve blind area coverage. Compared with corridor scenes, the traffic flow is busier and more complex, and occlusion is more likely to occur. The augment of such setting is of great necessity in intersections.
\end{itemize}
The detailed parameters are listed in Tab.~\ref{tab:sensors}

\begin{table}[bp]
  \small
  \centering
  \begin{tabular}{c|c|p{4.5cm}} 
  \Xhline{1.2pt}
    \multicolumn{2}{c|}{\textbf{Sensors}}      & \textbf{Details}  \\ \Xhline{1.2pt}
    Camera       & RGB                         & 42Hz, $1920\times1200$  \\ \hline
    \multirow{9}{*}{LiDAR}  & \multirow{3}{*}{80-beams} & 10Hz, $360^{\circ}$ horizontal FOV, $-25^{\circ}$ to $15^{\circ}$ vertical FOV, 1m to 230m capture range, $\pm 3cm$ error \\
    \cline{2-3}             & \multirow{3}{*}{32-beams} & 10Hz, $360^{\circ}$ horizontal FOV, $90^{\circ}$ vertical FOV, 0.1m to 30m capture range, $\pm 3cm$ error \\
    \cline{2-3}             & \multirow{3}{*}{MEMS} & 10Hz, $\pm 60^{\circ}$ horizontal FOV, $\pm 12.5^{\circ}$ vertical FOV, 0.5m to 200m capture range, $\pm 3cm$ error \\ \Xhline{1.2pt}
  \end{tabular}
  \caption{Sensor specifications in RCooper.}
  \label{tab:sensors}
\end{table}

\paragraph{Data Gathering}
The most representative 410 scenarios of 15 seconds duration are selected from a vast raw data pool. Sampling frequency is set as 3Hz, resulting in 30K frames of LiDAR point cloud (PC) and 50K frames of RGB images. Each frame of the corridor scene includes 4 RGB images and 2 pre-merged PCs (the PC of the multiline LiDARs group is pre-merged), and that of the intersection scene consists of 4 RGB images and 4 PCs (2 pre-merged PCs and 2 raw PCs of MEMES LiDARs). The synchronization between sensor agents is less than 50ms. 

\subsection{Coordinates and Data Annotation}

\paragraph{Coordinate System}
There are three different coordinate systems of our RCooper, i.e., the LiDAR coordinate system, the camera coordinate system, and the world coordinate system. The LiDAR coordinate system is regarded as the bridge, and we provide LiDAR-to-Camera and LiDAR-to-World calibration parameters for each frame. Besides, we annotate the 3D bounding boxes separately based on each infrastructure's LiDAR coordinate system such that each agent's sensor data alone can also be treated as independent roadside view perception tasks. The relative position between the two infrastructures in the same scene is mapped via the world coordinate system, and the system's origin is a virtual point of the local map.  

\paragraph{Labeling Approach}
We adopt a 3-steps labeling approach, including manual labeling for single infrastructure annotations, automatic labeling for cooperative annotations, and a final manual refinement step. We employ groups of professional annotators, and they exhaustively label each object in PC with the 7-degree-of-freedom 3D bounding box containing $x, y, z$ for centric location and $l, w, h, yaw$ for bounding box extent and orientation. There are ten semantic classes in total, belonging to five major classes, i.e., Vehicles (\textit{car, bus, truck, and huge\_vehicle}), Cyclists (\textit{bicycle, tricycle, and motorcycle}), Pedestrians, and Constructions (\textit{traffic\_sign and construction}). Each annotated object is assigned a unique object ID for the tracking task, and the object ID of the same object in one sequence is unique even when it is wholly occluded in some frames. To automatically generate the cooperative annotations based on the independent labels, we transform the objects from different LiDAR coordinates to the unified world coordinates and match the 3D bounding boxes via the Hungarian algorithm with Euclidean distance. We assign the same object ID for the matched objects and refine the bounding box according to all the corresponding annotations. If there is no matching object, it is introduced as a complement. Finally, we manually supervise and adjust the cooperative annotations and object IDs to obtain more accurate annotations. In addition, the whole dataset is desensitized before public release.

\subsection{Statistic and Scene Analysis}

\begin{figure}[t]
  \centering
  \includegraphics[scale=0.4]{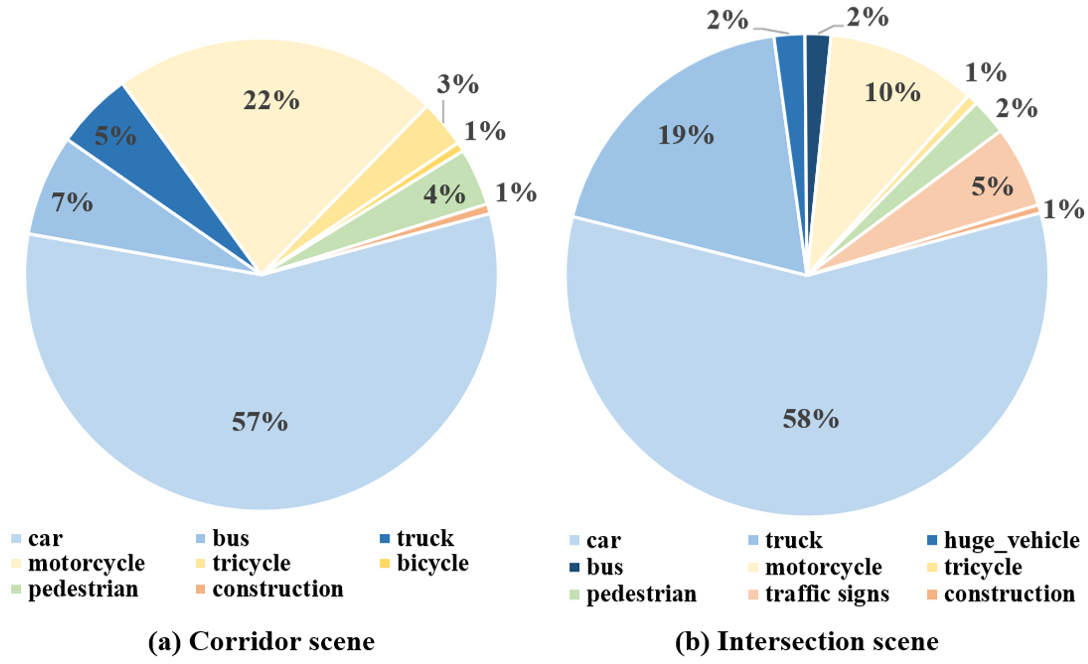}
  \caption{The distribution of semantic classes.}
  \label{fig:obj_stat}
\end{figure}

\paragraph{Data Statistics} 
It can be observed from Fig.~\ref{fig:obj_stat} that most of the objects (60\%) in RCooper belong to the Car class, and the other three vehicle classes take up 10\% in corridor scenes and 20\% in intersection scenes. The cyclists class ranks second, and the motorcycle has the most quantities. Since we focus more on vehicles than other road users, the ratio of pedestrians is limited. Besides, there are more construction labels in the intersection scene. 

\paragraph{Difference Analysis between Typical Scenes} 
There are two main differences between these two typical scenes:
\begin{itemize}
  \item \textbf{Spatial distribution of data} differs due to the topological characteristics. The corridor is long and narrow, while the intersection is spatial-wise centralized. As shown in Fig.~\ref{fig:ad-coop}(a.1), the cross-infrastructure data in corridor scenes is more clearly distributed in space, extending the sensing range and complementing the blind spot. Differently, the data from multiple infrastructures entwine each other and enhance the observation from various views in complex intersection scenes. Therefore, observation complementary is dominant in corridors, and the observation enhancement is dominant in intersections.
  \item \textbf{Data heterogeneity in intersection scenes} is the specific challenge in practical due to the hybrid-type of LiDAR systems as described in Fig.~\ref{fig:install}. Both multiline and MEMS LiDARs (with differing operating principles) are employed considering the construction cost, introducing severe data heterogeneity, which may result in failures of the existing cooperative perception methods (Tab.~\ref{tab:det-int}).
\end{itemize}

The abovementioned differences make our dataset an interesting but challenging playground. Not only the specialized approach for each scene, but also a unified approach for the entire roadside system, demand prompt solutions. 
\section{Task}
\label{sec:task}

Our dataset or its extension could further support multiple cooperative perception tasks, including detection, tracking, prediction, localization, counting, monitoring, etc. This paper focuses on cooperative detection and tracking tasks, i.e., integrating cross-infrastructure information to localize, recognize, and track 3D objects at a fixed traffic scene.

\subsection{Task Overview}

Theoretically, the task in this paper can be boiled down to two main sequential sub-tasks: roadside cooperative 3D detection and roadside 3D tracking. A straightforward collaborative approach is to achieve result-level fusion, i.e., late fusion, which merges multi-view detection results to the final detection results. The mainstream cooperative 3D detection framework employs early or intermediate fusion techniques, which first encode the roadside cooperative representation and employ a 3D detector to output the coordinates of 3D bounding boxes and object categories. Based on detected 3D bounding boxes, tracking by detection framework matches the objects and forms the trajectory ID. Two sub-tasks are illustrated in detail below.

\subsection{Roadside Cooperative Detection}

\paragraph{Task Description} 
The roadside cooperative detection task requires leveraging multiple LiDAR views to perform 3D object detection toward the corresponding district. Compared to single-view roadside detection, roadside cooperative detection has these challenges: data heterogeneity as mentioned before, and cooperative representation merits further enhancement. Our current benchmark does not aim to address these challenging issues absolutely but instead seeks to demonstrate the existence of these gaps and pave the way for subsequent research.

\paragraph{Input and Groundtruth} 
The input of roadside cooperative detection comprises multi-agent sequential frames and their relative posture. Take infrastructure node $i$ of a traffic region graph $\mathcal{G}$ as an example:
\begin{itemize}
  \item Sequential frames $\{C_{i}(t_{i}^{'})|t_{i}^{'} \leq T_{i}\}$ of node $i$ and $\{C_{N_{i}}(t_{N_{i}}^{'})|t_{N_{i}}^{'} \leq T_{N_{i}}\}$ of neighbors, where $T_i$ is the perception moment, $T_{N_{i}} \leq T_{i}$ is the capturing moments of neighbors, and $C(\cdot)$ denotes capturing function.
  \item Relative posture $M_i$ and $M_{N_i}$.
\end{itemize}

The perception outputs are the detected objects in the fixed traffic region, usually consisting of coordinates of detected 3D bounding boxes and the confidence score of the object category. Correspondingly, the groundtruth is the set of objects appearing in the region anytime and anywhere, which can be formulated as $GT = (GT_{i} \cup GT_{N_{i}}) \cap R$. $GT_i$ and $GT_{N_{i}}$ are groundtruth from node $i$ and neighbors, and $R$ is the interested region at fixed location. 

\paragraph{Benchmark Methodology} Most commonly adopted four fusion strategies are employed for roadside cooperative perception with state-of-the-art cooperative methods.

\begin{itemize}
  \item No Fusion: Only the single LiDAR point clouds are employed for detection, which is the baseline for comparing cooperative and non-cooperative methods.
  \item Late Fusion: 3D objects are detected for each LiDAR utilizing its sensor observations. Then, non-maximum suppression is adopted to merge and produce final outputs.
  \item Early Fusion: All the point clouds from multiple LiDARs are aggregated to form a more comprehensive point cloud, which can preserve complete information. Then, obey the no-fusion pipeline to generate detection results.
  \item Intermediate Fusion: The point clouds from each LiDAR are projected to a selected coordinate system and then are fed into neural feature extractors to encode intermediate features. Afterward, the encoded features are merged for cooperative feature fusion. Our benchmark employs several representative intermediate fusion methods, including AttFuse\cite{xu2022opv2v}, F-Cooper\cite{fcooper}, Where2Comm\cite{where2comm}, and CoBEVT\cite{cobevt}.
\end{itemize}

\paragraph{Evaluation Metrics} 
$400m \times 400m$ areas for fixed traffic scenes are chosen for perception evaluation. The common detection metric AP is used for 3D object detection evaluation. Specifically, AP values under different 3D IoU thresholds are reported for a more comprehensive evaluation.

\subsection{Roadside Cooperative Tracking}

\paragraph{Task Description}
The roadside cooperative tracking task is expected to show the superiority of the roadside cooperative temporal perception. There are two typical object tracking modes: joint detection and tracking, and tracking by detection. We concentrate on the latter in this paper.

\paragraph{Input and Groundtruth}
The input of roadside cooperative tracking is the roadside cooperative detection prediction, including coordinates of detected 3D bounding boxes and the confidence score of the object category. Moreover, the groundtruth is the correlation between trajectory IDs and object IDs. 

\paragraph{Benchmark Methodology}
We follow the previous works \cite{v2x-seq, v2v4real}, and implement AB3Dmot tracker \cite{9341164} in our benchmark. Based on the predictions from the cooperative detection models, the 3D Kalman filter and the Hungarian algorithm are employed by the AB3Dmot tracker to achieve efficient and high-quality tracking.

\paragraph{Evaluation Metrics}
The same evaluation metrics in \cite{nuscenes} and \cite{9341164} are adopted for roadside cooperative tracking evaluation, including 1) average multi-object tracking accuracy (AMOTA), 2) average multi-object tracking precision (AMOTP), and 3) scaled average multi-object tracking accuracy (sAMOTA), 4) multi-object tracking accuracy (MOTA), 5) mostly tracked trajectories (MT), and 6) mostly lost trajectories (ML).
\section{Benchmark Experiments}
\label{sec:benchmark}

In this section, we build two benchmarks respectively for roadside cooperative detection and tracking, expecting to provide effective and competitive benchmarks and pave the way for subsequent research.

\subsection{Implementation Details}

The dataset is split into the train/validation set as 4:1 (ratio of scenario), respectively. Consistent with \cite{v2v4real}, different categories are merged as the same class. For the RCooper detection task, PointPillar \cite{pointpillars} is adopted as the backbone of all models to extract features from the points cloud. They are trained for 50 epochs with a batch size of 16. The initial learning rate is set as $2\times 10^{-3}$ and is scheduled according to cosine annealing \cite{decoupled}. Adam optimizer \cite{adam} is adopted with a weight decay of $1\times10^{-4}$. For the tracking task, $F_{min}=1$ and $Age_{max}=2$ are set for the birth and death memory module according to the label criteria for the trajectory with shelters. In the data association module, we use $GIoU3D_{min}=-0.2$ as the threshold to filter the matching, consistent with the original AB3Dmot tracker \cite{9341164}.

\subsection{Roadside Cooperative Detection Results}

The roadside cooperative detection benchmark results for corridor scenes and intersection scenes are reported in Tab.~\ref{tab:det-cor} and Tab.~\ref{tab:det-int}, respectively. 

\begin{table}[t]
  \small
  \centering
  \begin{tabular}{c|ccc} 
  \Xhline{1.2pt}
    \textbf{Method}        & \textbf{AP@0.3}  & \textbf{AP@0.5}  & \textbf{AP@0.7} \\ \Xhline{1.2pt}
    No Fusion              & 40.0    & 29.2    & 11.1  \\ 
    Late fusion            & 44.5    & 29.9    & 10.8  \\
    Early fusion           & \bf 69.8    & 54.7    & 30.3 \\ \hline
    AttFuse \cite{xu2022opv2v}    & 62.7    & 51.6    & 32.1 \\
    F-Cooper \cite{fcooper} & 65.9    & 55.8    & 36.1 \\
    Where2Comm \cite{where2comm} & 67.1    & 55.6    & 34.3 \\
    CoBEVT \cite{cobevt}   & 67.6    & \bf 57.2    & \bf 36.2 \\ \Xhline{1.2pt}
    
  \end{tabular}
  \caption{Roadside cooperative detection benchmark of the corridor scene (\%).}
  \label{tab:det-cor}
\end{table}

\begin{table}[t]
  \small
  \centering
  \begin{tabular}{c|ccc} 
  \Xhline{1.2pt}
    \textbf{Method}        & \textbf{AP@0.3}  & \textbf{AP@0.5}  & \textbf{AP@0.7} \\ \Xhline{1.2pt}
    No Fusion              & 58.1    & 44.1    & 23.8  \\ 
    Late fusion            & \bf 65.1    & \bf 47.6    & 24.4  \\
    Early fusion           & 50.0    & 33.9    & 18.3 \\ \hline
    AttFuse \cite{xu2022opv2v}    & 45.5    & 40.9    & 27.9 \\
    F-Cooper \cite{fcooper} & 49.5    & 32.0    & 12.9 \\
    Where2Comm \cite{where2comm} & 50.5    & 42.2    & 29.9 \\
    CoBEVT \cite{cobevt}   & 53.5    & 45.6    & \bf 32.6 \\ \Xhline{1.2pt}
    
  \end{tabular}
  \caption{Roadside cooperative detection benchmark of the intersection scene (\%).}
  \label{tab:det-int}
\end{table}

\begin{figure}[t]
  \centering
  \includegraphics[scale=0.38]{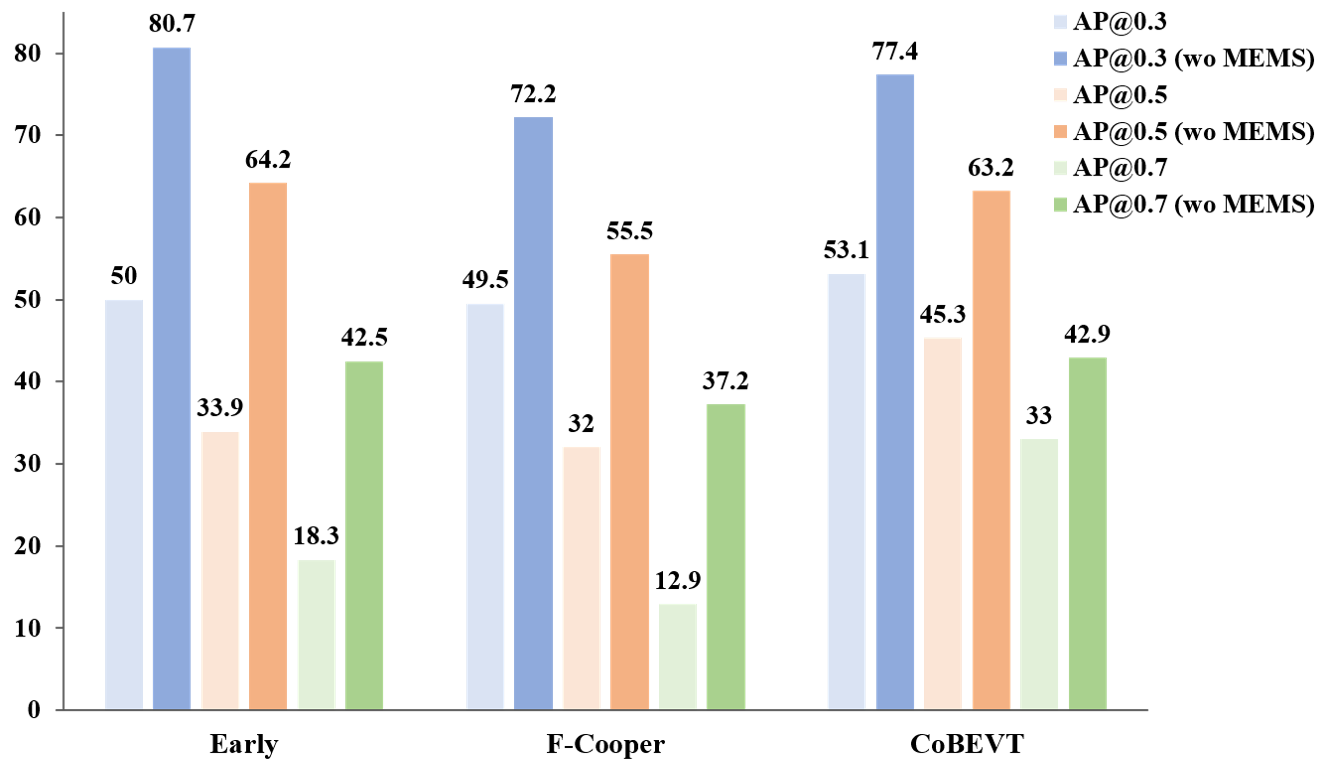}
  \caption{Influences of data heterogeneity. The performance of both early fusion and two representative feature fusion methods is boosted enormously when the two MEMS LiDARs are discarded, even if the number of collaborators decreases.}
  \label{fig:mems}
\end{figure}

\begin{table*}[t]
  \small
  \centering
  \begin{tabular}{c|cccccc} 
  \Xhline{1.2pt}
    \textbf{Method}        & \textbf{AMOTA $\uparrow$} & \textbf{AMOTP $\uparrow$} & \textbf{sAMOTA $\uparrow$} & \textbf{MOTA $\uparrow$} & \textbf{MT $\uparrow$} & \textbf{ML $\downarrow$} \\ \Xhline{1.2pt}
    No Fusion                     & 8.28 & 22.74 & 34.05 & 23.89 & 17.34 & 42.71 \\ 
    Late fusion                   & 9.60 & 25.77 & 35.64 & 24.75 & 24.37 & 42.96 \\
    Early fusion                  & \bf 23.78 & \bf 38.18 & 59.16 & 44.30 & \bf 53.02 & \bf 12.81 \\ \hline
    AttFuse \cite{xu2022opv2v}    & 21.75 & 35.31 & 57.43 & 44.50 & 45.73 & 22.86 \\
    F-Cooper \cite{fcooper}        & 22.47 & 35.54 & 58.49 & 45.94 & 47.74 & 22.11 \\
    Where2Comm \cite{where2comm}   & 22.55 & 36.21 & \bf 59.60 & 46.11 & 50.00 & 19.60 \\
    CoBEVT \cite{cobevt}          & 21.54 & 35.69 & 53.85 & \bf 47.32 & 47.24 & 18.09 \\ \Xhline{1.2pt}
    
  \end{tabular}
  \caption{Roadside cooperative detection benchmark of the corridor scene (\%).}
  \label{tab:trk-cor}
\end{table*}

\begin{table*}[t]
  \small
  \centering
  \begin{tabular}{c|cccccc} 
  \Xhline{1.2pt}
    \textbf{Method}        & \textbf{AMOTA $\uparrow$} & \textbf{AMOTP $\uparrow$} & \textbf{sAMOTA $\uparrow$} & \textbf{MOTA $\uparrow$} & \textbf{MT $\uparrow$} & \textbf{ML $\downarrow$} \\ \Xhline{1.2pt}
    No Fusion                     & 18.11 & 39.71 & 58.29 & 49.16 & 35.32 & 41.64 \\ 
    Late fusion                   & \bf 21.57 & 43.40 & \bf 63.02 & \bf 50.58 & \bf 42.75 & \bf 34.20 \\
    Early fusion                  & 21.38 & \bf 47.71 & 62.93 & 50.15 & 36.80 & 42.75 \\ \hline
    AttFuse \cite{xu2022opv2v}    & 11.84 & 36.63 & 46.92 & 39.32 & 29.00 & 53.90 \\
    F-Cooper \cite{fcooper}       & -4.86 & 14.71 & 0.00 & -45.66 & 11.52 & 50.56 \\
    Where2Comm \cite{where2comm}   & 14.21 & 38.48 & 50.97 & 42.27 & 29.00 & 45.72 \\
    CoBEVT \cite{cobevt}          & 14.82 & 38.71 & 49.04 & 44.67 & 33.83 & 35.69 \\ \Xhline{1.2pt}
    
  \end{tabular}
  \caption{Roadside cooperative tracking benchmark of the intersection scene (\%).}
  \label{tab:trk-int}
\end{table*}

It can be seen from Tab.~\ref{tab:det-cor} that all the cooperative approaches perform better than the no-fusion one, which aligns with our expectations. The cross-infrastructure cooperation in corridor scenes acts in a typical mode to achieve a comprehensive understanding of the long and narrow traffic area. Benefiting from the extension of the sensing range, the performance improvement is significant. The LiDAR-based CoBEVT \cite{cobevt} achieves the best performance in AP@0.5 and AP@0.7, while early fusion method achieves the best performance in AP@0.3. The corridor scenes provide a typical playground for cooperative detection research but from the roadside view.


Experimental results reported in Tab.~\ref{tab:det-int} show the detection performance in intersection scenes. Unlike the supplementary role in corridor scenes, cooperation at intersections is expected to learn a cooperative representation with observation from various views to understand the complex traffic scenario better. Note that the late-fusion achieves the best performance on $AP@0.3$ and $AP@0.5$ among all the cooperative methods, and the early fusion even performs worse than the no-fusion method. As for the intermediate fusion, CoBEVT \cite{cobevt} performs better on AP@0.7. Considering the data heterogeneity challenge in intersection scenes, the abovementioned phenomenon is unexpected but understandable.
\textbf{\textit{Firstly}}, the impact of data heterogeneity is more evident for early fusion since the simple integration (without fine-grained design) of data cannot cope with the heterogeneity problem and even makes the data distribution more complex to model, resulting in worse performance compared with the no-fusion method. \textbf{\textit{Secondly}}, late fusion theoretically overcomes the problem via the result-level fusion with bounding boxes, and the performance advantages demonstrate the effectiveness of cooperative perception at intersections. The intermediate fusion approach can deal with the challenge to some extent by feature-level cooperation, and the advantages of cooperative representation lead to a better performance at a higher IoU threshold, i.e., $AP@0.7$. To further explore the impact of data heterogeneity, a data-level ablation study is reported in Fig.~\ref{fig:mems}. The excluding of point clouds from MEMS LiDAR directly makes the cooperative approaches act in a typical manner. Therefore, further research can utilize RCooper to study how to fully leverage the sensing data and overcome the heterogeneity challenge in practical scenes. Referring to multimodal learning techs, we think a possible way lies in encoding heterogeneous data into a unified feature space by feature extraction incorporating distribution consistency constraints, such as aligning distribution via KL divergence.

Apart from the failure of SOTA methods in intersection scenes, some SOTA methods are not as effective as simple fusion method (early or late) in both scenes. Another reason may lie in the scenario gap. Some SOTA Methods designed for vehicle-centric scenarios, which can leverage vigorous development in vehicle-side perception, struggle with infrastructure-specific challenges (e.g., larger variations of mounting heights and pitch angles compared with vehicle-side) \cite{rope3d, bevheight}. Specific methods for roadside cooperative perception deserve further investigation.

\subsection{Roadside Cooperative Tracking Results}

The roadside cooperative tracking benchmark results for corridor scenes and intersection scenes are reported in Tab.~\ref{tab:trk-cor} and Tab.~\ref{tab:trk-int}, respectively. 

For corridor scenes, the cooperative tracking performance is better than that of the no-fusion method, demonstrating the effectiveness of roadside cooperative temporal perception, as shown in Tab.~\ref{tab:trk-cor}. Compared methods present fierce competition: early fusion method achieves the best performance in AMOTA, AMOTP, MT, and ML metrics, Where2Comm \cite{where2comm} achieves the best performance in sAMOTA metric, while CoBEVT \cite{cobevt} achieves the best performance in MOTA metric.



For intersection scenes, the late-fusion strategy outperforms others, as shown in Tab.~\ref{tab:trk-int}. Since the detection predictions are affected by the data heterogeneity, the experimental results of AB3Dmot present a similar pattern as the detection. Besides, the tracking results also depend on the temporal-wise continuity of detection predictions, so it cannot generate satisfactory trajectories if the instance is not detected stably in adjacent frames, which results in the worse performance of F-Cooper (whose MT value reduces to $11.52\%$). The tracking-by-detection strategy is susceptible to the detection performance. How to learn a roadside cooperative representation for the tracking task and how to leverage the spatial-temporal contexts in an end-to-end manner in the roadside scenes need further exploration.

\section{Conclusion}
\label{sec:conclusion}

To extend the boundaries of both autonomous driving and traffic management, \textbf{R}oadside \textbf{Coo}perative \textbf{Per}ception \textbf{(RCooper)}, a real-world, large-scale dataset, is released in this paper, which is expected to boost roadside cooperative perception towards round-the-clock area-coverage perception for a restricted traffic area. Our dataset consists of 30K manually annotated sensor data groups (images and point-cloud) covering two typical traffic scenes (i.e., intersection and corridor). A competitive benchmark is constructed to pave the way for subsequent research, and the experimental results demonstrate the effectiveness of roadside cooperation perception and reveal the direction of future work.

\paragraph{Limitation.} 
In this paper, we release the RCooper and build the corresponding benchmarks with several representative approaches. Although it opens the gate, the Garden of Eden still deserves further exploration. As the aforementioned discussion, foreseeable future works include:  1) learning a unified roadside cooperative representation to not only overcome the data heterogeneity challenges but also leverage spatial-temporal multimodal sensor data; 2) exploring a unified approach for end-to-end perception and other tasks; 3) finding a solution for practical challenges, like calibration noises, leveraging the advantages of cross-infrastructure cooperation.
{
    \small
    \bibliographystyle{ieeenat_fullname}
    \bibliography{ref}
}

\clearpage
\setcounter{page}{1}
\setcounter{section}{0}
\renewcommand\thesection{\Alph{section}}
\maketitlesupplementary

\section{Appendix}
\label{sec:appendix}
This supplementary document is organized as follows:
\begin{itemize}
  \item Additional information and visualization of RCooper is shown in Appendix \ref{sec:vis_dataset}.
  \item The visualization of roadside cooperative perception results is supplemented in Appendix \ref{sec:vis_perres}, offering an illustrative overview of the performance evaluation of our established benchmark.
  \item Some preliminary camera-based experiment results is shown in Appendix \ref{sec:camera}.
\end{itemize}

\section{Dataset Visualization}
\label{sec:vis_dataset}
\begin{figure*}[htbp]
  \centering  \includegraphics[scale=0.42]{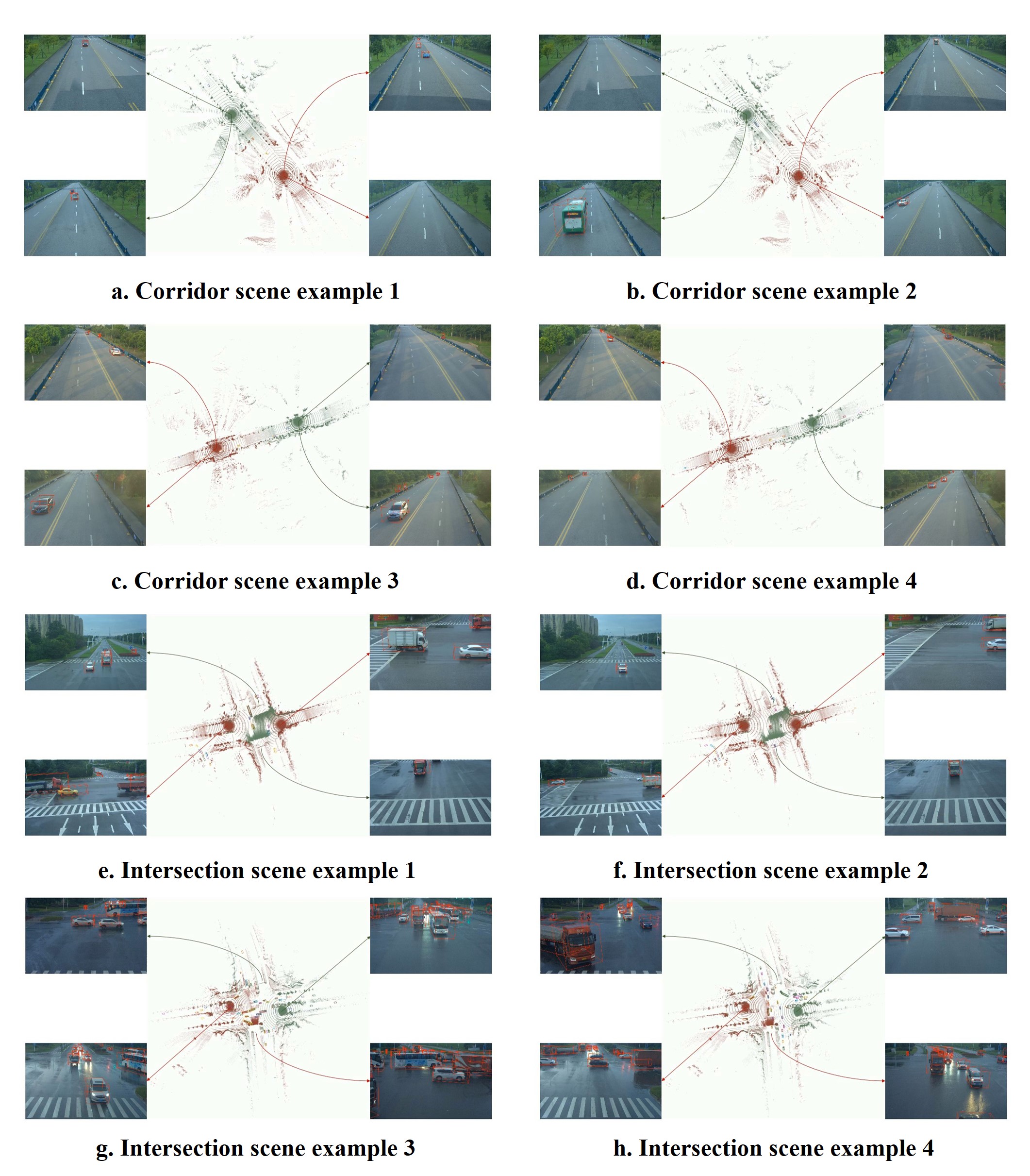}
  \caption{Visualization Examples of the RCooper Dataset.}
  \label{fig:dataset_demo}
\end{figure*}

More visualizations of the proposed RCooper dataset are shown in Fig~\ref{fig:dataset_demo}. These visualizations serve as a valuable supplementary to the discussion in Section 3.1, where two representative traffic scenes (i.e., intersection and corridor) are presented. Each figure within this context comprises an aggregated 3D LiDAR point cloud image and four camera images captured from distinct views. For the corridor scenes, the point clouds generated by two different LiDAR sensors are depicted in red and green, respectively. The infrastructure-side cooperation can effectively extend the sensing range to cover the whole corridor scene. For the intersection scenes, the point clouds generated by multiline LiDARs and MEMES LiDARs are denoted in red and green, respectively. The observations from multiple views serve to mitigate the challenges posed by occlusions in intricate intersection scenarios.

\section{Perception Results Visualization}
\label{sec:vis_perres}
More visualizations of benchmark results are shown in this section. 

\begin{figure*}[htbp]
  \centering  \includegraphics[scale=0.38]{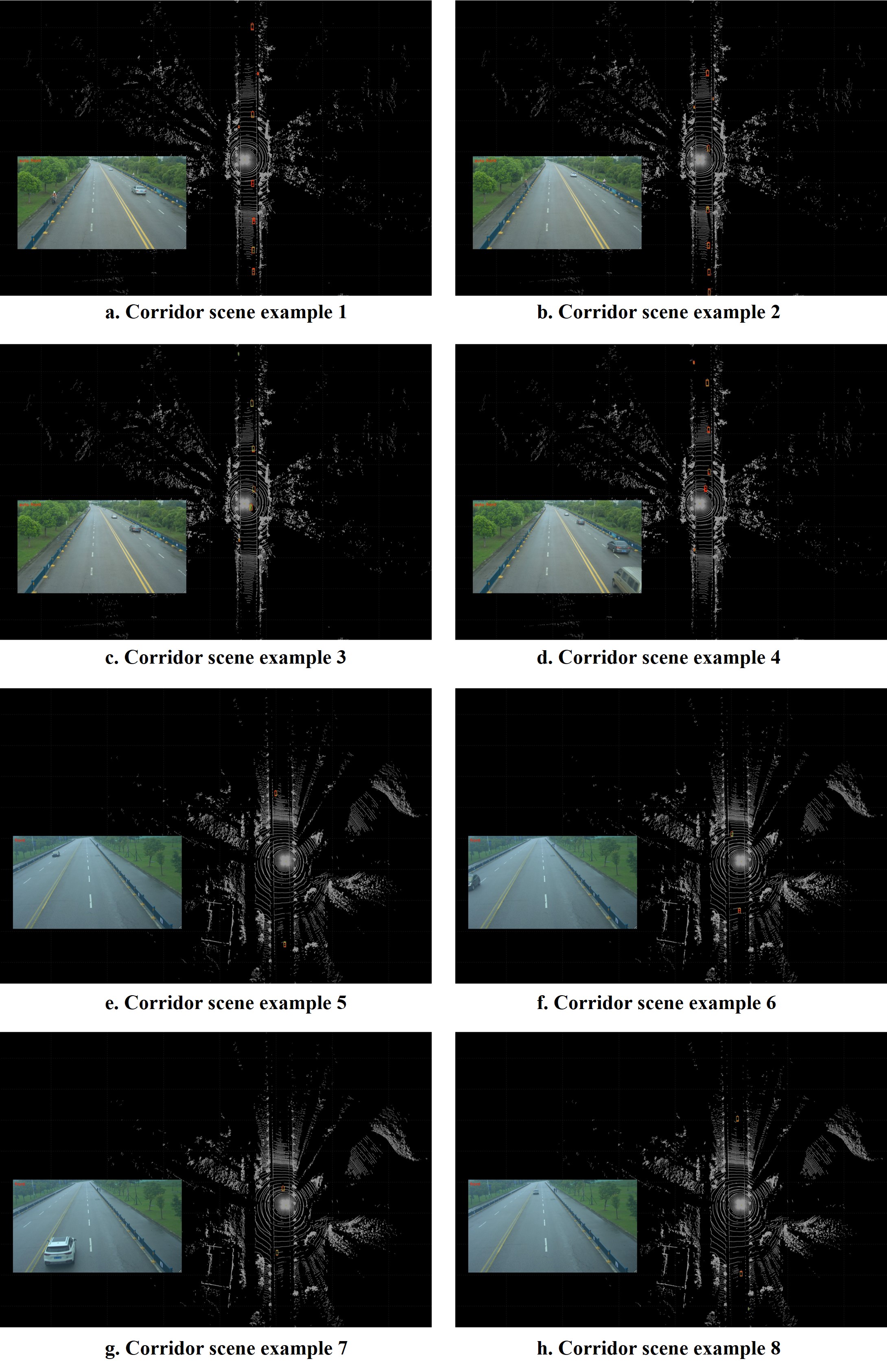}
  \caption{Visualization Examples of the Perception Results for Corridor Scenes. Red and yellow bounding boxes represent the prediction and ground truth, respectively.}  \label{fig:corridor_perception_demo}
\end{figure*}

\begin{figure*}[htbp]
  \centering  \includegraphics[scale=0.38]{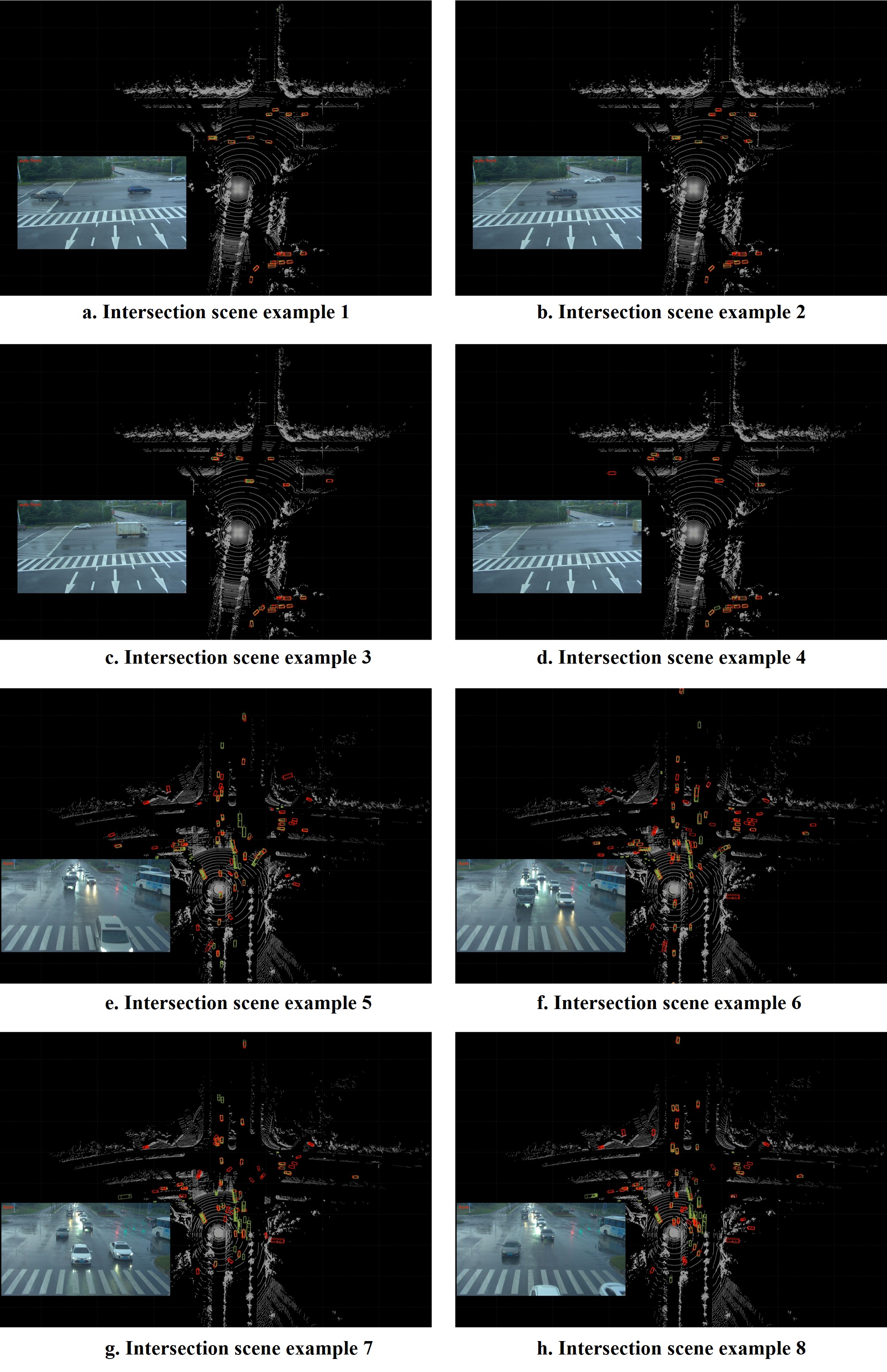}
  \caption{Visualization Examples of the Perception Results for Intersection Scenes. Red and yellow bounding boxes represent the prediction and ground truth, respectively.}  \label{fig:intersection_perception_demo}
\end{figure*}

Typical perception benchmark results for corridor scenes (referred to as CoBEVT results) and intersection scenes (referred to as late fusion results) are illustrated in Fig~\ref{fig:corridor_perception_demo} and Fig~\ref{fig:intersection_perception_demo}, respectively. Within each figure, alongside the LiDAR points and a camera image, we have delineated the predicted objects with red outlines and the ground truth objects with yellow outlines. For the corridor scenes (in Fig~\ref{fig:corridor_perception_demo}) and uncrowded intersection scenes (in Fig~\ref{fig:intersection_perception_demo}. a, b, c and d), the perception process proves to be highly effective. For the crowded intersection scenes (in Fig~\ref{fig:intersection_perception_demo}. e, f, g and h), the presence of occlusions and the inherent complexity of the scenarios introduce false negatives and false positives into the perception process. These challenges merit further investigation in future research endeavors.

\section{Preliminary Camera-based Experiment Results}
\label{sec:camera}
We provide some preliminary camera-based experiment results in this section. A single-view camera suffers from the inherent blind spot, which can be theoretically mitigated by roadside cooperative perception. Camera-based late-fusion results in both corridor and intersection scenes are reported in Tab.~\ref{tab:det-camera}. It's worth noting that the data annotation of our datasets is based on point cloud data, and meanwhile the sensing range of camera group is smaller than that of LiDAR group, which means that the label may need filtering on demand considering camera group's view.

\begin{table}[htbp]
  \small
  \centering
  \begin{tabular}{c|cccc} 
  \Xhline{1.2pt}
    \textbf{Scenario}     & \textbf{3D@0.3}  & \textbf{3D@0.5}  & \textbf{BEV@0.3}    & \textbf{BEV@0.5}\\ 
    \Xhline{1.2pt}
    Corridor              & 26.99  & 3.53  & 38.70 & 16.15 \\ 
    Intersection          & 11.32  & 2.41  & 12.19 & 7.89 \\
    \Xhline{1.2pt}
  \end{tabular}
  \vspace{-0.1cm}
  \caption{Camera-based cooperative detection AP metric (\%).}
  \label{tab:det-camera}
\end{table}
\end{CJK}
\end{document}